%% file: acl_latex.tex
\pdfoutput=1

\documentclass[11pt]{article}

\usepackage[]{acl}

\usepackage{times}
\usepackage{latexsym}

\usepackage[T1]{fontenc}

\usepackage[utf8]{inputenc}

\usepackage{microtype}

\usepackage{inconsolata}
\usepackage{amsfonts}
\usepackage{amsmath}
\usepackage{amssymb}
\usepackage{mathtools}
\usepackage{enumitem}
\usepackage{graphicx}
\usepackage{lipsum, babel}
\usepackage{booktabs}
\usepackage{multirow}

\title{How much do contextualized representations encode long-range context?}

\author{Simeng Sun\hspace{1.5em}Cheng-Ping Hsieh\\
NVIDIA\\
\texttt{\{simengs,chsieh\}@nvidia.com}
}

\begin{document}
\maketitle

\input{000_abstract}

\input{1_introduction}

\input{2_methodology}

\input{3_exp_setup}
\input{4_case_transformer}

\input{5_case_other_models}

\input{6_related_work}

\input{7_conclusion}

\bibliography{anthology,custom,colm_bib}

\appendix
\input{0_appendix}

\end{document}

%% file: 000_abstract.tex
\begin{abstract}
We analyze contextual representations in neural autoregressive language models, emphasizing long-range contexts that span several thousand tokens. Our methodology employs a perturbation setup and the metric \emph{Anisotropy-Calibrated Cosine Similarity}, to capture the degree of contextualization of long-range patterns from the perspective of representation geometry. We begin the analysis with a case study on standard decoder-only Transformers, demonstrating that similar perplexity can exhibit markedly different downstream task performance, which can be explained by the difference in contextualization of long-range content. Next, we extend the analysis to other models, covering recent novel architectural designs and various training configurations. 
The representation-level results illustrate a reduced capacity for high-complexity (i.e., less compressible) sequences across architectures, and that fully recurrent models rely heavily on local context, whereas hybrid models more effectively encode the entire sequence structure. Finally, preliminary analysis of model size and training configurations on the encoding of long-range context suggest potential directions for improving existing language models.
\end{abstract}

%% file: 1_introduction.tex
\section{Introduction}

In neural autoregressive language models~\citep{Mikolov2010RecurrentNN,lmlimit,gpt2}, each token is predicted from a contextual representation, created by integrating relevant information from the sequence history into the previous token's representation. This process, which we refer to as ``context-mixing'', is enabled by architectural innovations like attention~\citep{attn15,transformer} and linear recurrence~\citep{mamba}.
Existing analyses of contextualized representations primarily focus on short sequences of tens to hundreds of tokens~\citep{ethayarajh-2019-contextual}, whereas modern language models handle hundreds of thousands of tokens in a single context window. In this work, we study contextualized representations conditioned on long-range contexts, defined as spanning at least a few thousand tokens.

We analyze contextualized representations by examining their responses to perturbations applied to long-range contexts, similar to prior analyses~\citep{khandelwal-etal-2018-sharp,sun-etal-2021-long}. To quantify these changes, we adopt \emph{anisotropy-calibrated cosine similarity} (ACCS), a metric directly adapted from the analysis by~\citet{ethayarajh-2019-contextual}. ACCS computes the cosine similarity between a representation and its counterpart conditioned on the perturbed context devoid of original context structures. To enforce fair comparison across layers and models, the similarity is calibrated by anisotropy -- the expected cosine similarity estimated over a large sample. Intuitively, ACCS measures the degree of contextualizing long-range context patterns. The lower the ACCS score, the more contextualized the representations.

To provide a primer of our experimental setup and ACCS, we begin with a case study on a standard decoder-only Transformer. The case model utilizes Rotary Position Embedding~\citep[][RoPE]{rope}, whose hyperparameter $\theta$ effectively influences context scaling~\citep{xiong-etal-2024-effective,ropescalinglaw}. We generate several dozen model instances by varying $\theta$ to examine the relationship between perplexity, downstream task performance, and ACCS. As expected, representations become increasingly more contextualized, as manifested in the decreasing ACCS with layer depth. Results also indicate that models with similar perplexity can correspond to markedly different downstream task performance, which can potentially be explained by the extent to which hidden representations are contextualized by long-range content, as reflected by the ACCS scores.

Next, we extend the analysis to other models. To address implementation transferability~\citep{narang-etal-2021-transformer}, we pre-train six 0.5B models covering multiple architectures within the same framework on OpenWebText~\citep{gpt2}, including Transformers using different positional encoding methods~\citep{rope,alibi}, recurrent models~\citep[][mLSTM]{xlstm}~\citep[][Mamba-2]{mamba2}, as well as hybrid models~\citep[][Griffin]{griffin}~\citep[][HybridMamba]{hybridmamba}. Additionally, we analyze four large open-access models from the \texttt{llama3} and \texttt{llama3.1} series~\citep{dubey2024llama3herdmodels}. 

As anisotropy serves as a crucial baseline for computing ACCS, we first take a closer look at this key component while controlling for confounding factors such as token frequency. One observation is that representations tend to become less isotropic, or clustering into increasingly narrow subspaces, as the context becomes less compressible, i.e., deficient in regularities. We then proceed with ACCS and demonstrate that: (1) Fully recurrent models and Transformers using ALiBi~\citep{alibi} position encoding rely primarily on local context to predict future tokens. (2) In contrast, RoPE-based transformers are over-contextualized by noises in distant context for unseen sequence lengths. (3) Hybrid models exhibit better contextualization of the entire context, neither over-relying on local context nor failing to distinguish distant signals from noise. (4) Both architectural design and training configurations affect a model’s ability to recognize patterns as sequence length increases, with hybrid models and aligned open-access models generally performing better.

In summary, we present the first analysis of contextualized representations with regard to long-range context and the effect of architectural design in a controlled setup. We hope our analysis will shed light on the future development of more effective long-context language models.

%% file: 2_methodology.tex
\section{Methodology}\label{sec:method}

\paragraph{Notations} We denote a sequence with
\[
\mathbf{S} \coloneqq \underbrace{x_1, x_2, \dots, x_T}_{\texttt{prefix}}, \underbrace{ y_1, y_2, \dots, y_N}_{\texttt{suffix}}.
\] 
$\mathbf{S}$ is partitioned into a prefix and a suffix, with the prefix being much longer than the suffix ($T \gg N$). This setting lets us study cases where suffix tokens are provided with a sufficiently long context, similar to sliding window perplexity evaluation~\citep{baevski2018adaptive}, which scores tokens at the end of a context window. We focus on the contextualized (or hidden, or intermediate) representation at layer $L$ of a language model:
\[
    \mathbf{h}^{(L)}_{y_i} = \phi_L\bigg(\mathbf{h}^{(L-1)}_{x_1}, \mathbf{h}^{(L-1)}_{x_2},\dots \mathbf{h}^{(L-1)}_{y_{i}}\bigg),  \
\]
\noindent where $\phi_L$ is a context-mixing operator at layer $L$, such as attention~\citep{attn15} or other recent novel designs~\citep{mamba}. The context-mixing operator passes necessary information from the history to the representation of $y_i$, and is trained with other components to ensure low surprisal of the ground-truth next token.

\paragraph{Perturbation \& self-similarity.} To study how much a suffix token is contextualized by long-range prefix, we apply perturbation~\citep{khandelwal-etal-2018-sharp,sun-etal-2021-long} operations $\xi(\cdot)$ to the prefix string and observe corresponding changes in the suffix token representations. We employ \emph{cosine similarity} to quantify the change in high-dimensional space.\footnote{We provide other metrics (dot product and condition number of sample covariance matrix) in Appendix~\ref{sec:appendix_other_metrics}. }  For simplicity, let $\mathbf{h}$ and $\mathbf{h'}$ represent the contextualized representations of the same suffix token $y_i$ produced by the same model at layer $L$, where $\mathbf{h'}$ is conditioned on a perturbed prefix. That is, $\mathbf{h'}$ corresponds to the contextual representation of suffix token in the sequence
\[
\mathbf{\Tilde{S}} \coloneqq \underbrace{\xi(x_1, x_2, \dots, x_T)}_{\texttt{perturbed prefix}}, \underbrace{ y_1, y_2, \dots, y_N}_{\texttt{suffix}}.
\]
\noindent 
We compute ``self-similarity'' by averaging over $m$ pairs of suffix tokens and their counterparts given perturbed prefixes:
\[
 \text{self}\_\text{similarity} (\mathbf{h}, \mathbf{h'}) = \frac{1}{m}\sum_{i=1}^m \frac{\langle \mathbf{h}_i, \mathbf{h}_i' \rangle}{\Vert \mathbf{h}_i \Vert \cdot \Vert \mathbf{h}_i'  \Vert}
\]
The self-similarity differs from that by~\citet{ethayarajh-2019-contextual} in that we compute at the corpus level, against representations induced by perturbations, and do not enforce constraints on token types. A self-similarity value close to 1 indicates that the applied perturbation does not significantly alter the direction of the representation, implying limited contextualization of the perturbed range.

\paragraph{Calibration with Anisotropy.} Calibrating cosine similarity with a baseline is crucial for fair comparisons across different architectures or different layers of a same model, which often show drastically different angular dispersion in latent space. A self-similarity of 0.95 suggests a strong correlation between $\mathbf{h}$ and the perturbed $\mathbf{h'}$ when the suffix representations are on average weakly correlated (e.g., 0.3) to each other. However, it suggests dissimilarity when representations are highly correlated with each other (e.g., 0.99). As such, we compute anisotropy-calibrated cosine similarity (ACCS). The anisotropy baseline is the expected pairwise cosine similarity over the representations sampled from the distribution $\mathcal{D}_{hh'}$, which contains representations given both the original and the perturbed prefix:
\[
\mathcal{A} = \mathbb{E}\text{ }_{\substack{\mathbf{h}_i,\mathbf{h}_j \sim \mathcal{D}_{hh'}\\i\neq j} }\big[ \text{cos}(\mathbf{h}_i, \mathbf{h}_j)\big].
\]
\noindent Anisotropy reflects how concentrated the representations are direction-wise in latent space, with 0 indicating maximal dispersion and 1 maximal concentration without utilizing the rest directions of the latent space. 
Anisotropy-calibrated cosine similarity (ACCS) is thus:
\[
\text{ACCS} = \text{self}\_\text{similarity} (\mathbf{h}, \mathbf{h'}) - \mathcal{A}.
\]
\noindent High ACCS suggests limited contextualization of the perturbed context range. Geometrically, high ACCS occurs when perturbation minimally alters the representation while all representations are dispersed maximally in angular measure. By extension, a low ACCS score indicates greater contextualization of the perturbed context range.

%% file: 3_exp_setup.tex
\section{Experimental Setup} \label{sec:experiment}

\paragraph{Models} To address concerns on implementation transferability~\citep{narang-etal-2021-transformer}, we re-implement recently proposed architectures within the framework while referencing open-source repositories. Specifically, we pre-train models of $\sim$0.5B parameters on OpenWebText~\citep{gpt2} with context length 1024 and with equal number of optimization iterations. A summary of the models is presented in Table~\ref{tab:models}.\footnote{Details regarding hyperparameters and performance on standardized benchmarks can be found in Appendix~\ref{sec:appendix_perf}
and~\ref{sec:appendix_hparams}.} In addition to the smaller models, we evaluate larger open-access models, including \texttt{llama3-8b-base}, which is pre-trained with context length of 8K, and models post-trained with context length of 128K: \texttt{llama3.1-8b-base}, \texttt{llama3.1-8b-instruct}, and \texttt{llama3.1-70b-base}. Overall, we evaluate a diverse array of models that vary in architectural design, model size, and training configurations.
\input{tables/models}

\paragraph{Residual stream} Modern deep learning models incorporate residual connections~\citep{He2015DeepRL} to stabilize training. Specifically, the residual is added back to the output of a sublayer, which can be either a context-mixing operator or a feed-forward (MLP) layer. 
The residual stream is our primary focus as it reveals how representations evolve across layers. Given similar trend observed in preliminary experiments, the following analyses present only the residual stream representations after the context-mixing modules.

\paragraph{PG19 and synthetic sequences.} We evaluate all models on the PG19~\citep{Rae2020Compressive} test set, which contains 100 public domain books, as well as entirely out-of-distribution synthetic token sequences. The latter is inspired by the view of language models as general pattern machines~\citep{mirchandani2023large} and lossless compressors~\citep{deletang2024language,huang2024compression}. We generate synthetic inputs by injecting regularities into otherwise ``incompressible'' strings produced by uniformly sampling from a large vocabulary. The regularity we inject is the simplest possible one -- the periodic repetition of a string.

\paragraph{Shuffling as perturbation.} The perturbation applied for computing self-similarity is the shuffling operation, which disrupts any structural patterns in the prefix while preserving the entropy with respect to the sequence unigram distribution. Therefore, the self-similarity measures how much precise structure the model is able to ``memorize'' beyond the fuzzy semantic field determined by token frequency distribution. We compute self-similarity over $\sim$100K pairs of suffix tokens, and estimate the anisotropy baseline using 500M token pairs. The random seed is controlled so that all architectures process the same perturbed prefixes for the same set of suffix tokens.

%% file: tables/models.tex
\begin{table}[!t]
    \centering
    \scalebox{0.75}{
    \begin{tabular}{l|l}
    \hline
       \textbf{Model Name}  & \textbf{Model Type} \\ \hline
       GPT+RoPE~\citep{rope}  & Transformer\\
       GPT+ALiBi~\citep{alibi} & Transformer\\
       Mamba-2~\citep{mamba2} & Recurrent \\
       mLSTM~\citep{xlstm} & Recurrent \\
       Griffin~\cite{griffin} & Hybrid\\
       HybridMamba~\citep{hybridmamba} & Hybrid\\ \hline
    \end{tabular}}
    \caption{Summary of small models we pre-train from scratch on OpenWebText~\citep{gpt2}.}
    \label{tab:models}
\end{table}

%% file: 4_case_transformer.tex
\section{Case study: standard GPT with varying RoPE base frequency} \label{sec:case_transformer}


\paragraph{Transformer \& Rotary Position Embedding} Recent Transformer-based language models often adopt Rotary Position Embedding~\citep[][RoPE]{rope} to inject the positional information. Roughly speaking, RoPE encodes the relative distance between two token positions (query and key tokens), and is designed to decay (though not smoothly) the dependencies between the tokens as their distance increases. One hyperparameter that influences the decay rate is the base $\theta$, with larger $\theta$ generally allowing for a slower decay of the inter-token dependencies, thereby enabling a Transformer to ``look'' farther. We refer readers to other materials~\citep{rope,ropescalinglaw} for more formal understanding of RoPE.

\paragraph{Layerwise evolution of representations in GPT+RoPE.} ACCS measures the degree of contextualizing long-range structural patterns in our defined experimental setup, as progressing over more context-mixing modules, one should expect more contextualized representations (lower ACCS score). We verify this in two settings: PG19 data at 1K tokens (the max pre-training length) and synthetic sequences at 16K tokens. While trends in self-similarity and anisotropy differ, Figure~\ref{fig:rope_layerwise} confirms that ACCS decreases with layer depth, aligning with intuition. The difference between these settings stems from anisotropy. In the synthetic setting, GPT+RoPE exhibits extremely high anisotropy (and in consequence, high self-similarity), suggesting that representations occupy a much narrower cone in latent space in deeper layers, despite diverse synthetic patterns in the prefix. In section~\ref{sec:look_at_anisotropy}, we take a closer look at anisotropy by relating it to sequence complexity.

\input{figures/gptrope_layerwise}

\paragraph{Perplexity vs. downstream task performance vs. ACCS.} Language models are typically evaluated using intrinsic metrics (e.g., perplexity) and extrinsic metrics (e.g., downstream task performance), yet the relationship between the two remains inconclusive~\citep{lu2024controlledstudylongcontext,gao2024trainlongcontextlanguagemodels}. Here, we present an empirical analysis showing perplexity is not a reliable indicator of the downstream task performance. This can be potentially explained by the fact that the same perplexity can be achieved when representations are contextualized to various degrees by long-range prefix, as evidenced by different ACCS scores. To obtain a sufficient number of models for studying the relationship, we gradually increase the RoPE $\theta$, which is a naïve way of extending context length without further fine-tuning. We evaluate the models with more than 40 different $\theta$ values at context size of 4K. The metrics we study are as follows: 
\begin{itemize}[leftmargin=*]
    \item \textbf{Suffix perplexity (intrinsic)}: The perplexity of suffix tokens on PG19. Akin to perplexity evaluated with sliding window approach. 
    \vspace{-0.2em}
    \footnote{We provide non-overlapping perplexity in Appendix~\ref{sec:appendix_perf}.}
    \item \textbf{RULER S-NIAH accuracy (extrinsic)}: We adopt three simple retrieval tasks from a synthetic  benchmark~\citep[][RULER]{hsieh2024ruler} and compute the average accuracy over context sizes in \{1K, 2K, 4K, 8K\}.
    \vspace{-0.2em}
    \item \textbf{ACCS (representation level)}: We compute ACCS on PG19 by perturbing the distant long-range prefix (first half of prefix). For simplicity, we show results of the last layer. 
\end{itemize}
Figure~\ref{fig:rope_base} shows that RoPE base $\theta$ effectively alters the suffix perplexity, and that the same suffix perplexity can correspond to vastly different downstream task performances. The bottom plot shows that RoPE base also affects long-range contextualization. Specifically, we observe a clear phase transition as $\theta$ increases: initially, the representations exhibit reduced reliance on distant prefix (as indicated by the rising ACCS values), which suggests that the drop in perplexity is largely driven by better modeling of local context. However, as $\theta$ continues to increase, ACCS begins to decline again, indicating that tokens are becoming more influenced by distant context again, accompanied by increasing perplexity. This intuitively makes sense as stronger contextualization of distant prefix may not be always beneficial, especially due to the presence of irrelevant noises and the inherent locality of natural language~\citep{futrell-2019-information}. 
\input{figures/rope_base}

%% file: figures/gptrope_layerwise.tex
\begin{figure}
    \centering
    \includegraphics[width=\linewidth]{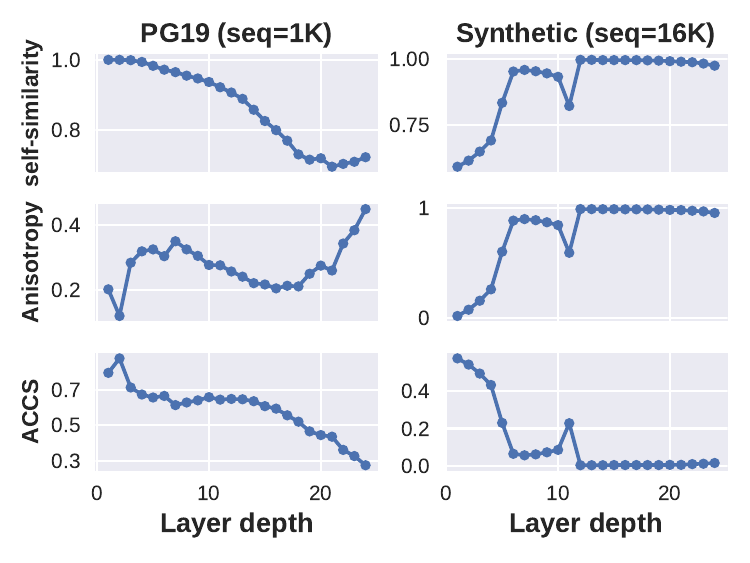}
    \caption{Layerwise evolution of contextualized representations. We evaluate two settings that differ primarily in their anisotropy (expected cosine similarity), with the synthetic setting showing highly correlated representations, and consequently high self-similarity, despite diverse synthetic patterns in the prefix. Regardless of the input, representations become increasingly more contextualized by long-range prefix, as shown in the decreasing trend of ACCS.
    }
    \label{fig:rope_layerwise}
\end{figure}

%% file: figures/rope_base.tex
\begin{figure}
    \centering
    \includegraphics[width=\linewidth]{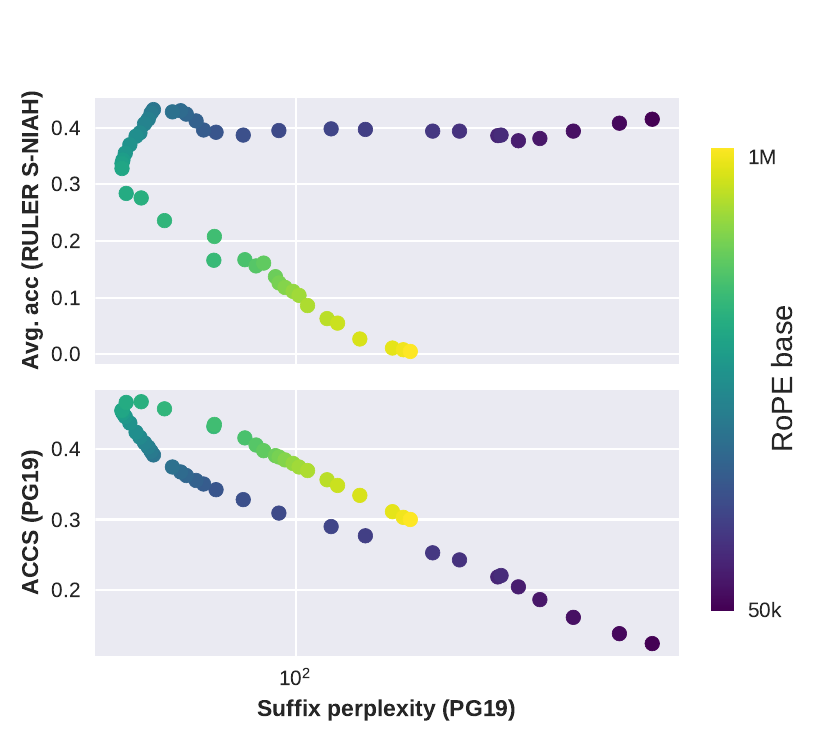}
    \caption{Relationship between suffix perplexity, downstream task performance, and ACCS. Same perplexity can be reached when representations are contextualized by distant context to various degrees (measured by ACCS) and when the downstream task performance differs significantly.
    }
    \label{fig:rope_base}
\end{figure}

%% file: 5_case_other_models.tex
\input{figures/anisotropy_vs_models}

\section{Investigating modern language models} \label{sec:case_others}

In this section, we extend the analysis to nine other language models, including various architectural designs (Table~\ref{tab:models}) and large open-access models.

\subsection{What causes difference in anistropy $\mathcal{A}$?} \label{sec:look_at_anisotropy}

The large difference in anisotropy between PG19 and synthetic setting (\S~\ref{sec:case_transformer}) motivates us to take a closer look at the calibration baseline in ACCS. Recent work~\citep{godey-etal-2024-anisotropy} suggests that anisotropic representations are inherent to self-attention, while other research~\citep{su2023contrastive} finds it only in English-only models. In this section, we examine anisotropy from a different angle, showing that when sequence token distribution is strictly controlled, models tend to exhibit higher anisotropy as the sequence becomes less compressible.

\paragraph{Controlled uni-gram distribution.} First, we control the set of suffix tokens to eliminate the effect of token distribution on anisotropy. When models are compared on tokens from different distributions (e.g., different languages), high anisotropy can be the result of more skewed token distributions (e.g., dominance of certain frequent tokens). Next, we verify that prefixes contain diverse patterns. Since we compute anisotropy at corpus level, contextual patterns that repetitively appear in different examples can also lead to high anisotropy. Our setup using PG19 and synthetic sequences naturally restricts the suffix token distribution while ensuring prefix diversity.

\paragraph{Sequence Complexity.} When token frequency is controlled, what other factors can affect anisotropy? Inspired by the view of language models as lossless compressors~\citep{rae23,deletang2024language}, we relate anisotropy to how compressible a sequence is, or the sequence complexity. A sequence is of high complexity when the shortest description is at least the length of itself (i.e., incompressible). In contrast, human languages are highly compressible due to the presence of regularities. Since the complexity of a sequence is relative to the compressor,\footnote{A high complexity sequence can be compressible when a compressor is trained to explicitly reduce it.} to make it a static property that we can compare different models, we use an non-neural compressor to measure the complexity. Specifically, we compute the average \emph{prefix compression rate} (i.e., compressed prefix size / raw prefix size, in bytes) using LZMA, a variation of~\citet{lemplelziv}.

\paragraph{Models show anisotropic representations when prefixes are less compressible.} The shuffling perturbation disrupts existing regularities (e.g., common $n$-grams or hierarchical dependencies), making the sequences less compressible. To demonstrate a trend, we divide the prefix into chunks and gradually increase the local shuffling window size, similar to the setup by~\citet{kallini-etal-2024-mission}.  Figure~\ref{fig:aniso_vs_mdl} shows that regardless of architecture, model size, or training configuration, the anisotropy of contextualized suffix tokens increases. Similar trend is also observed when the natural language patterns are not disturbed, as presented in Appendix~\ref{sec:appendix_anisotropy}, where we divide PG-19 into bins based on prefix compression rate. These observations indicate reduced angular dispersion in latent space, or increased correlation in directions as compression rate increases. An immediate implication is that models lose representational power in angular measure as input complexity increases, though this effect is less pronounced in larger models (e.g., \texttt{llama3.1-70b-base}), which has a larger model dimension.

\subsection{How much is long-range context encoded?}

Anisotropy-calibrated cosine similarity (ACCS) measures the (dis)similarity of a representation to itself when the prefix is perturbed. Therefore, by adjusting the position and range of the perturbation, one can evaluate how much a hidden representation is contextualized by different ranges of a prefix. To understand the contextualization of the entire prefix, we begin the perturbation from the beginning of the prefix and gradually extend the right boundary towards the suffix tokens. 

\input{figures/accs_vs_models}

\paragraph{GPT+RoPE without context extension overly contextualizes prefix.} Both the small GPT+RoPE model and \texttt{llama3-8b-base}, when tested on sequences longer than their pre-training lengths, exhibit lower ACCS scores and display trends that differ significantly from other models. As discussed in \S~\ref{sec:case_transformer}, low ACCS\footnote{We provide the anisotropy and self-similarity scores in appendices. The low ACCS is mainly driven by higher anisotropy compared to the rest models.} is not always desirable -- encoding noises in the long-range prefix can lead to meaningless representations.  While most models rely heavily on local context (as shown by the sharp drop in ACCS when perturbations are close to the suffix in Figure~\ref{fig:accs_vs_mdl}), GPT+RoPE shows uniform contextualization of the prefix, regardless of the proximity of applied perturbations to the suffix. Similarly, \texttt{llama3-8b-base} does not show strong biases toward immediate local prefixes, though it does contextualize the distant content less.

\paragraph{Hybrid models show better long-range contextualization.} Alternative context-mixing operators, when combined with attention mechanisms, have been shown to perform well on long-context tasks such as retrieval from prefix~\citep{jamba15}. Our analysis provides a representation-level explanation for the advantage of hybrid models. Recurrent models, like Mamba-2 and mLSTM, exhibit nearly flat ACCS curves in Figure~\ref{fig:accs_vs_mdl} (left) when the right boundary of the perturbation is far from the suffix tokens. This indicates a strong reliance on short-range context for predicting future tokens while leveraging limited long-range contextual patterns. We observe similar trend in GPT using ALiBi positional encoding, which functions like a soft sliding window that gradually decays distant signals. In contrast, hybrid models, especially HybridMamba,  effectively contextualize the entire sequence, as indicated by the gradually decreasing ACCS without plateauing. These models also show relatively low perplexity reported in Appendix~\ref{sec:appendix_perf}, unlike exploding perplexity with GPT+RoPE. This suggests that hybrid models encode information from the full context more effectively, without over-reliance on local context when predicting the next token.

\paragraph{Open-access models show strong reliance on local context.} Figure~\ref{fig:accs_vs_mdl} (right) shows that llama3.1 series display similar trends as attention-free models and GPT+ALiBi: suffix representations are weakly contextualized by distant prefix and are dominated by local context. We conjecture that models have seen similar or identical sequences during training, therefore model weights serve as ``additional context'' and enable good prediction without integrating global patterns. To avoid the impact of parametric knowledge, we use synthetic out-of-distribution sequences in the following experiment.

\input{figures/accs_vs_models_varlen}
\subsection{How do contextualized representations change with context size?}

\paragraph{Fixed language source.}  Unlike the fixed-length input in the previous section, we shift to entirely synthetic variable-length sequences, where the ``regularity'' presented in a given sequence can be fully controlled. Specifically, starting from fully random sequences (i.e., incompressible), we inject regularity by periodically repeating an $L$-token string with stride $k$ and $L > k$ ($L=200$, $k=56$). A prefix with such regularities becomes increasingly more ``compressible'' as the prefix length increases, and thus increasingly more dissimilar to the perturbed version where regularities are disrupted. In this experiment, we shuffle the entire prefix to understand the effect of context spanning long-range.

\paragraph{Effect of architectural design.}  Figure~\ref{fig:accs_vs_mdl_varlen} (left) displays two distinct ACCS trends when prefix length increases. Both hybrid models and attention-based model exhibit a gradual decrease in ACCS, indicating they gradually ``discern'' the regularity as context length increases. In contrast, fully recurrent models plateau up to 4K tokens, beyond which ACCS scores drop, suggesting that recurrent models may require sufficient accumulation of a pattern in the long-range prefix to reflect it in the contextual representation geometry.

\paragraph{Observations on Llama models.} The open-access Llama models belong to the same family, differing only in size or training configurations. While they generally follow similar trends, the aligned model \texttt{llama3.1-8b-instruct} shows a lower ACCS, indicating it encodes prefix patterns into geometry more than the other models. Interestingly, the 70b model exhibits less contextualization at shorter context lengths but eventually catches up with the smaller models as the sequence length increases. This raises the question of whether an optimal model size exists for a given sequence length.

%% file: figures/anisotropy_vs_models.tex
\begin{figure*}
    \centering
    \includegraphics[width=0.44\linewidth]{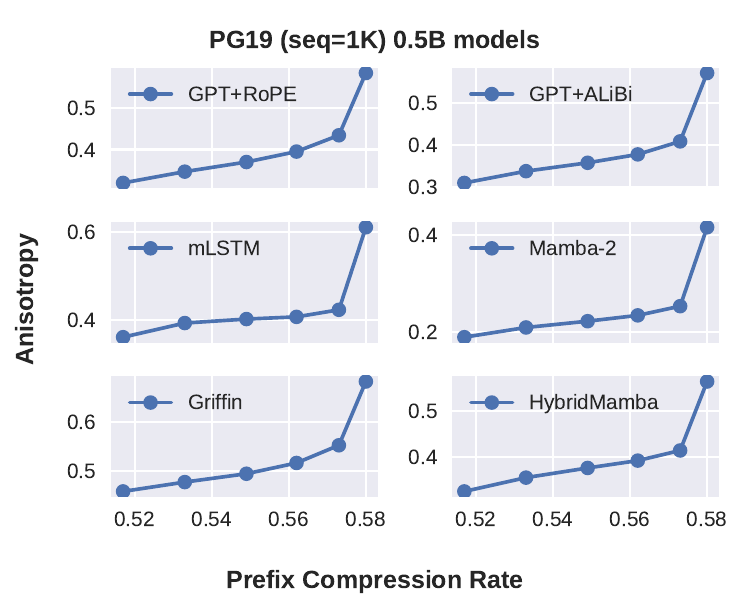}\includegraphics[width=0.44\linewidth]{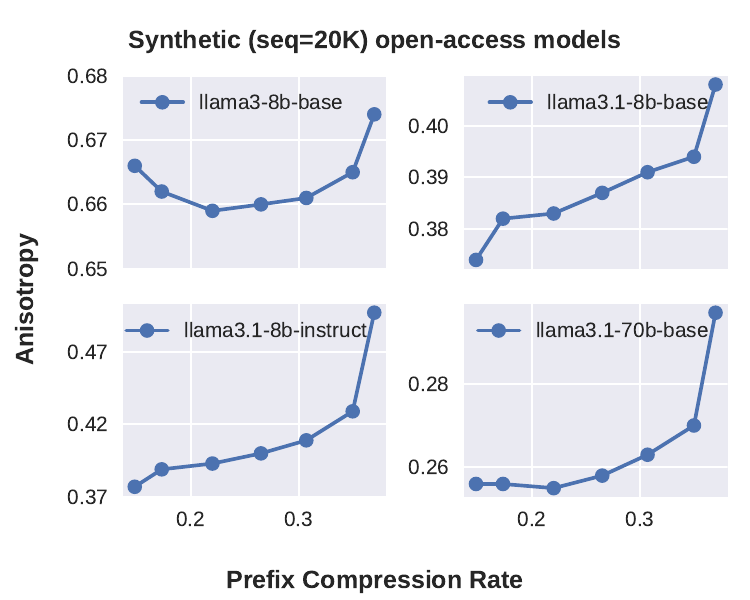}
    \caption{Models exhibit increasingly anisotropic representations as prefixes become less compressible, or have high compression rate (i.e., compressed prefix size / raw prefix size, using LZMA compression).}
    \label{fig:aniso_vs_mdl}
\end{figure*}

%% file: figures/accs_vs_models.tex
\begin{figure*}
    \centering
    \includegraphics[width=0.46\linewidth]{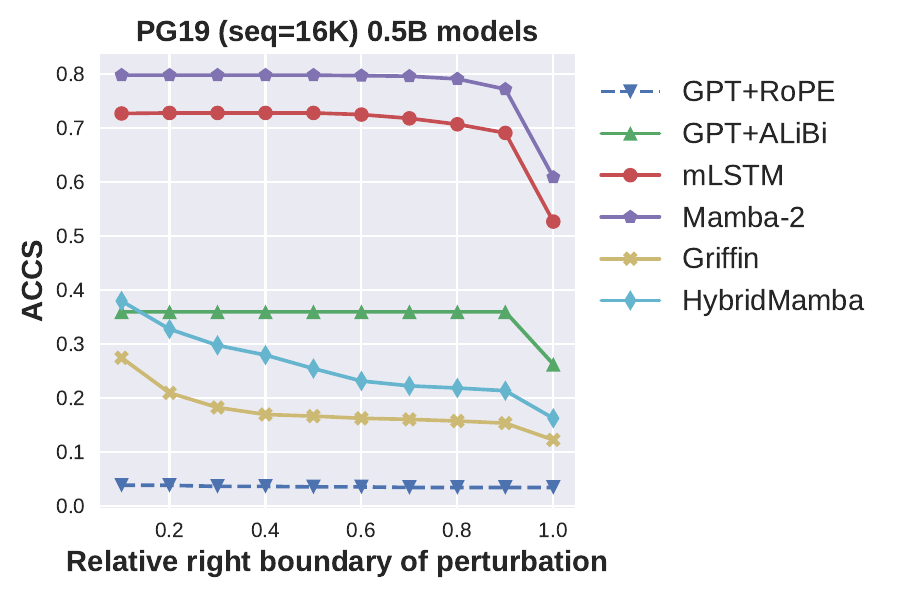}\includegraphics[width=0.46\linewidth]{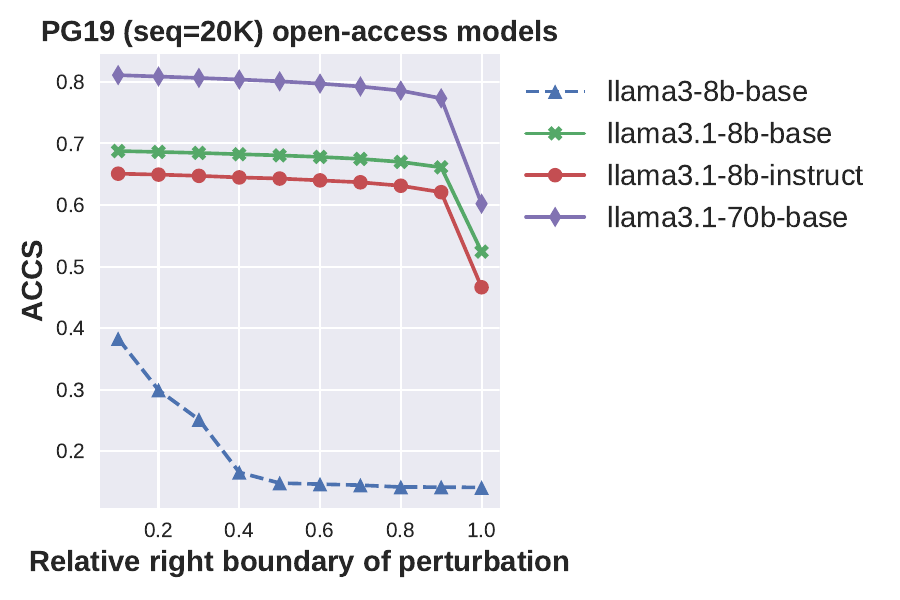}
    \caption{We apply perturbations from the beginning of the prefix and gradually extend the right boundary towards suffix tokens (relative boundary = 1.0). RoPE-based Transformers (dashed lines) display low ACCS when perturbing the majority or all of the prefix, likely due to over-contextualization of noises in the prefix. Fully recurrent models (mLSTM, Mamba-2) and GPT with ALiBi demonstrate sudden drops in ACCS when perturbing nearby tokens, indicating stronger reliance on short-range context while minimally contextualized by distant prefix (plateau on the left). In contrast, hybrid models demonstrate a continuous downward trend, indicating more effective contextualization of the entire prefix.}
    \label{fig:accs_vs_mdl}
\end{figure*}

%% file: figures/accs_vs_models_varlen.tex
\begin{figure*}
    \centering
    \includegraphics[width=0.44\linewidth]{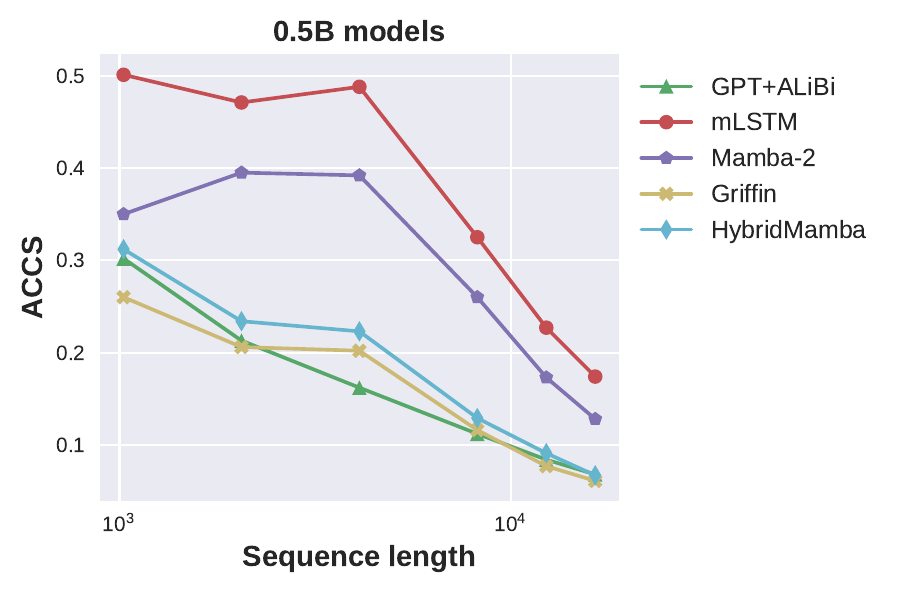}\includegraphics[width=0.44\linewidth]{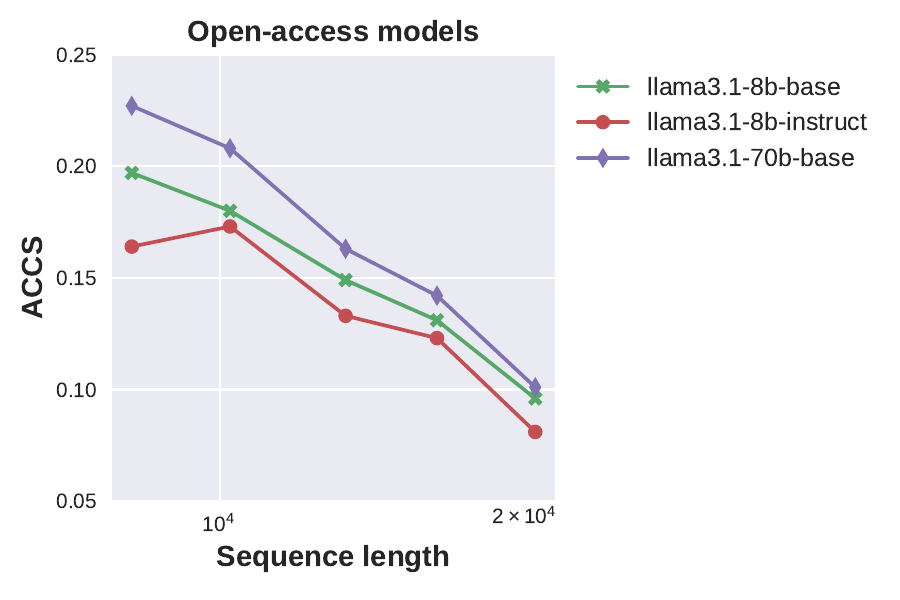}
    \caption{We evaluate models on synthetic sequences with fully controlled patterns that become increasingly recognizable as sequence length grows. All models show increased contextualization of regularities, though fully recurrent models need some accumulation of patterns (initial flat lines). Interestingly, the larger 70b model encodes less prefix patterns at shorter sequence lengths but catches up with smaller models with larger context length. }
    \label{fig:accs_vs_mdl_varlen}
\end{figure*}

%% file: 6_related_work.tex
\section{Related Work}

Contextualized representations have been shown to encode useful linguistic features~\citep{liu-etal-2019-linguistic,hewitt-manning-2019-structural} examined with linear probing. Recent work~\citep{morris-etal-2023-text} studies these representations by inverting them back into short sequences. In contrast, this work focuses on long inputs and employs a different methodology by analyzing the representation geometry, building on prior analyses~\citep{ethayarajh-2019-contextual,cai2021isotropy}. Perturbations are often employed to analyze the effect of context~\citep{khandelwal-etal-2018-sharp,sun-etal-2021-long} and the robustness in adversarial scenarios~\citep{li-etal-2021-contextualized}; more controlled causal interventions 
 are used to study other phenomena such as verbatim memorization in language models~\citep{huang2024demystifyingverbatimmemorizationlarge}. 

Long-range context often touches the length extrapolation regime. Better length generalization~\citep{Lake2017GeneralizationWS,anil2022exploringlengthgeneralizationlarge,deletang2023neural,zhou2023algorithmstransformerslearnstudy} performance is often driven by novel designs, such as modifications to position encoding~\citep{pi,sun-etal-2023-length,xiong-etal-2024-effective,jin2024llmmaybelonglmselfextend}, attention~\citep{wu2022memorizing,wang2023augmenting}, adaptive layer depth~\citep{fan2024loopedtransformerslengthgeneralization} in Transformer and completely new architectures~\citep{mamba,hyena,retnet,peng-etal-2023-rwkv,ma2024megalodonefficientllmpretraining,deltanet}, which deserve additional investigation. Many of the designs can be considered as applying ``regularization'' to the global long-range context, and are relevant to~\citet{Rosenfeld1996AME}'s max entropy approach mixing local and global context~\citep{bau-andreas-2021-neural}.
Recent novel designs are often evaluated using perplexity and downstream tasks. Recent work~\citep{lu2024controlledstudylongcontext} demonstrates a linear relationship between perplexity and long-context downstream tasks. Our results suggest a more intricate relationship between these two. 

Anisotropic representations have been observed by prior works~\citep{gao2018representation} and were shown to be alleviated by contrastive method~\citep{contrastivegen,gao-etal-2021-simcse}, spectrum control~\citep{Wang2020Improving}, and proper regularization~\citep{zhang-etal-2020-revisiting}. Recent work suggests that anisotropy is inherent to self-attention~\citep{godey-etal-2024-anisotropy}, and it is greatly affected by token frequency~\citep{zhou2021frequencybaseddistortionscontextualizedword,puccetti-etal-2022-outlier}. While anisotropic representations are often considered harmful, recent work shows it depends on the downstream task~\citep{ait-saada-nadif-2023-anisotropy}. We study anisotropy as an intermediate step towards understanding contextualized representations, by relating anisotropy to an intrinsic property of the input string, inspired by the view of language modeling as compression~\citep{deletang2024language,gu2024optimallearninglanguagemodels}. Finally, our analysis suggests better contextualization of the entire long-range context in hybrid models, echoing recent positive results from hybridization of attention and other layer modules~\citep{jamba15,samba,hybridmamba}.

%% file: 7_conclusion.tex
\section{Conclusion}
\vspace{-0.4em}
We presented an analysis of contextual representations in neural autoregressive language models, with a focus on long-range context. We quantified the impact of long-range patterns with a perturbation experiment setup and the metric \emph{anisotropy-calibrated cosine similarity}. A simple case study of standard GPT demonstrated that similar perplexity can be reached when representations encode long-range patterns to various degrees, which further manifested in different downstream task performances. We then extended the analysis to other architectures, revealing a connection between sequence complexity (i.e., compressibility of a sequence) and anisotropy. Finally, through representation-level results, orthogonal to intrinsic and extrinsic evaluation, we showed the benefits of large model size, hybridizing attention with alternative modules, and potentially aligning base models.

\section*{Limitations}
The presented analysis is conducted over ten language models given long-form narratives and synthetic sequences. It can further consolidate some of the findings by 1) examining other more recent architectures~\citep{deltanet,ttt} and open-access models, 2) increasing the pre-trained model sizes, 3) evaluating on other domains or even modalities. In the presented experiments, the perturbation operation was constrained to simple token shuffling. It can be interesting to investigate other perturbation operations, such as shuffling while maintaining tri-gram or word-level statistics, disrupting only the hierarchical dependencies or syntactic patterns, more involved token replacement without changing plug-in entropy, etc. These additional perturbations can provide new insights into how contextual representations encode the said features. Our experiments on synthetic sequences were limited to the simplest possible regularity added to a close-to-random string. Other regularities can be explored and help inspect model generalization behavior. Finally, our analysis primarily focused on base models with limited investigation on aligned models and zero exploration on more practical scenarios, e.g., downstream tasks involving reasoning. Analyzing the representations given many-shot demonstrations, which contain repetitive patterns, can be an interesting future direction to expand the presented analysis.

\section*{Acknowledgement}
We thank Ilya Loshchilov for tuning our Transformer baseline. We thank both Ilya Loshchilov and Boris Ginsburg for numerous helpful discussions.

%% file: 0_appendix.tex
\clearpage
\onecolumn

\section{Expected cosine similarity, dot product, and condition number} \label{sec:appendix_other_metrics}
\input{figures/other_metrics}

\section{Model hyperparameters \& other details} \label{sec:appendix_hparams}
\input{tables/hparams}
\noindent The hyperparameters used by GPT+RoPE and GPT+ALiBi are shown in Table~\ref{tab:hparams}. For ALiBi, we use the default slopes specified in the NeMo~\citep{kuchaiev2019nemotoolkitbuildingai} framework.\footnote{\url{https://github.com/NVIDIA/NeMo/blob/main/nemo/collections/nlp/modules/common/megatron/position_embedding/alibi_relative_position_embedding.py}} All architectures share the same vocab size and training data. The models we pre-trained do not tie the input and output embeddings. For other architectures, we modify the model depth to achieve roughly same number of parameters ($\sim$0.47B) across all models. When re-implementing alternative architectures, we find it crucial to cast data into torch.float32 during context mixing (e.g. cumsum of forget gates in mLSTM) for stabilized training. For mLSTM we use conv1d kernel size 4, qkv number of heads 4, projection factor 2. For HybridMamba, we mix 8\% attention layers with 62\% of Mamba-2 layers and 30\% MLP layers. For Griffin, we insert local attention layers (window size = 1024) every two layers of RG-LRU layers. For both hybrid models, the first layer is consistently a non-attention layer. We anneal to final learning rate of 0. The peak learning rate for Griffin and mLSTM is 0.001, with the rest of parameters the same as specified in Table~\ref{tab:hparams}.

\section{Model performance} \label{sec:appendix_perf}
We adapt the evaluation framework lm-evaluation-harness~\citep{eval-harness} for both intrinsic and extrinsic evaluation. For all models, we use greedy decoding except for mLSTM, which we use top\_k=2, top\_p=0.6 due to repetitions in the greedy decoding output. A summary of model performance on standardized benchmark can be found in Table~\ref{tab:downstream_Task}. We find that models that exhibit strong recency bias also achieve non-zero results for long-context retrieval tasks at 8K context size, 8 times the max training length. We empirically find that these models perform better when the context is highly compressible or when relevant information is close to the query.
\input{tables/downstream_task}

\section{Additional discussion on Anisotropy and sequence complexity} \label{sec:appendix_anisotropy}
\paragraph{Anisotropy without disrupting natural language distribution} In section~\ref{sec:look_at_anisotropy}, we have demonstrated that anisotropy increases as the prefix becomes less compressible, which is achieved by disrupting natural language distribution via shuffling tokens. Here, we present complementary results showing the trend also exists when language patterns are not disrupted by binning examples based on their prefix compression rate. We omit open-access models as very likely they have been trained on PG-19. Figure~\ref{fig:pg19_bin} shows similar trend with a smaller range of prefix compression rate.
\input{figures/pg19_bin_scatter}

\paragraph{Less compressible prefix leads to worse in-context retrieval accuracy.} In Figure~\ref{fig:niah_llama31}, we plot the retrieval accuracy of \texttt{llama3.1-8b-base} with increasing number of distractor needles in the ``haystack'' of highly compressible repetitive sentences. While it remains unclear in prior works whether increased anisotropy hurts downstream task performance or not, we show here that the increased anisotropy, which happens alongside the increase in prefix compression rate, co-occurs with decreased in-context retrieval performance. The decreased downstream task performance is also consistent with prior work~\citep{machina-mercer-2024-anisotropy}. 
\input{figures/niah_llama31}

\paragraph{Two modes of collapse.} We have shown that as the prefix becomes less compressible, the contextualized representations become increasingly aligned in certain directions. But what do these  representations collapse into? Our preliminary analysis identifies two distinct modes: (1) the representation corresponding to a uniform distribution over the vocabulary, and (2) the representation corresponding to the unigram prior of the pre-training corpus. Figure~\ref{fig:collapse_repr} displays two cases where representations are anisotropic: (left) sequences that exceed the maximum training length, especially for GPT+RoPE without training-free extension, and (right) sequences that are less compressible due to the absence of patterns. Consider a sequence of length $L$ consisting of tokens from a finite vocabulary of size $|V|$. Increasing $L$ leads to exponential increase of high complexity strings for which models demonstrate decreased representational capacity, as reflected by the increase in anisotropy, or deterioration into predicting unigram prior. The reduced isotropy can potentially explain common observations of repetitive output of frequent words (e.g. ``aaaaa'', ``the the'') given long prefixes, while more evidence is needed.  Precisely quantifying the ``regularization'' required for the expanding prefix string can potentially help guide better design of new language models.  
\input{figures/collapse_representation}

\section{Numerical results}
We detail the numbers for plots in section~\ref{sec:case_transformer} and section~\ref{sec:case_others} with Table~\ref{tab:rope_base_table},~\ref{tab:accs_pg19_slm},~\ref{tab:accs_pg19_llama},~\ref{tab:accs_synth_slm}, and~\ref{tab:accs_synth_llama}.
\input{tables/accs_pg19_slm}
\input{tables/accs_pg19_llama}
\input{tables/accs_synth_slm}
\input{tables/accs_synth_llama}
\input{tables/rope_base_table}

%% file: figures/other_metrics.tex
\begin{figure*}[!h]
    \centering
    \includegraphics[width=0.49\linewidth]{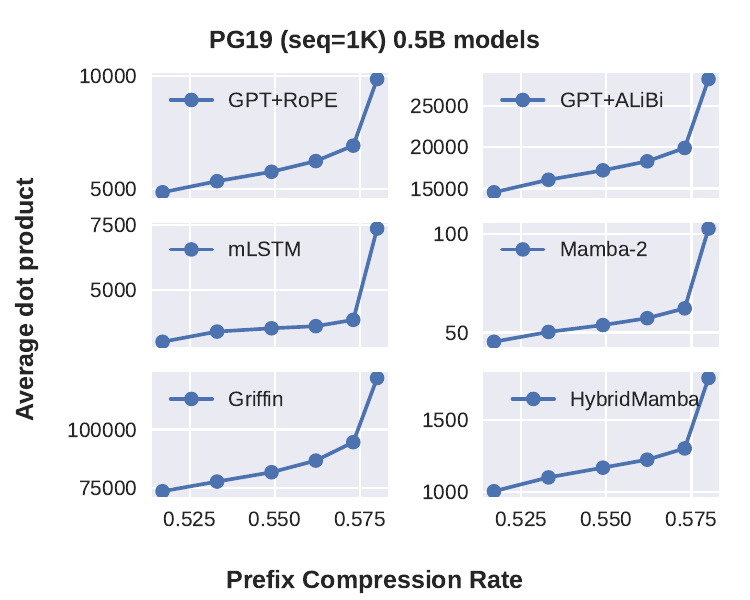}\includegraphics[width=0.49\linewidth]{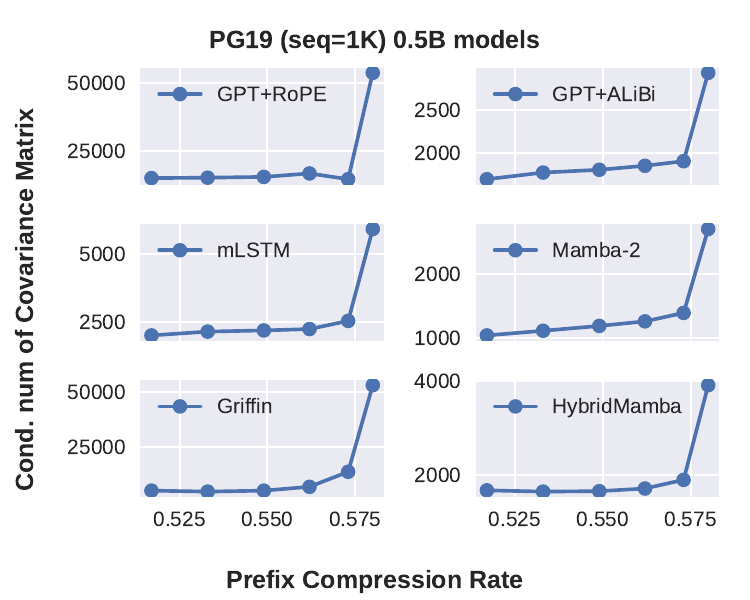}
    \caption{Recent work~\citep{steck2024} pointed out that cosine-based metric may not be reliable. We provide additional metrics that reflect the geometry of hidden representations. The two additional metrics (dot product and condition number of sample covariance matrix) take additional information into account (e.g., magnitude) and demonstrate similar increasing trend as the average pairwise cosine similarity, which we used as the main metric for evaluating anisotropy. }
    \label{fig:other_metrics}
\end{figure*}

%% file: tables/hparams.tex
\begin{table}[!h]
    \centering
    \begin{tabular}{llll}
    \toprule
    \textbf{Configuration}    &  \textbf{Value} & \textbf{Configuration}    &  \textbf{Value}\\ \midrule
    Num. Layers & 24 & Steps & 100K\\
    Num. attention heads & 16 & Normalization & RMSNorm\\
    Rotary base & 10000 & Activation & GeGLU \\
    Model (embedding) dimension   &  1024 & Optimizer & AdamW \\
    feed-forward hidden dimension & 4096 & Weight decay & 0.1\\
    Vocab size & 32000 & Betas & (0.9, 0.95)\\
    Tokenizer type & sentencepiece & Warmup steps & 2000\\
    Training context length & 1024 & Init method std & 0.02 \\
    Global batch size & 512 & Learning rate & 0.003\\
    \bottomrule
    \end{tabular}
    \caption{Configurations for training GPT+RoPE. Architecture-specific hyper-parameters are described in Appendix~\ref{sec:appendix_hparams}.}
    \label{tab:hparams}
\end{table}

%% file: tables/downstream_task.tex
\begin{table*}[!h]
\scalebox{0.82}{
\begin{tabular}{@{}lcccccccccc@{}}
\toprule
\multirow{2}{*}{Model} & wsc273 & hellaswag & \multicolumn{2}{c}{ARC\_e} & LAMBADA & \begin{tabular}[c]{@{}c@{}}Wikitext\\ Len=1024\end{tabular} & \begin{tabular}[c]{@{}c@{}}PG19\\ Len=16384\end{tabular} & \multicolumn{3}{c}{\begin{tabular}[c]{@{}c@{}}RULER acc.\\ (S-NIAH)\end{tabular}} \\ \cmidrule(l){2-11} 
       & acc    & acc\_norm & acc& acc (fs=5)    & acc     & word ppl    & token suffix ppl & 1K & 4K& 8K       \\ \midrule
GPT+RoPE       & 63.0 & 41.8    & 50.3     & 52.4& 45.7  & 25.9& 5554.3   & 98.9       & 0 & 0\\
GPT+ALiBi      & 63.7 & 44.5    & 51.6     & 54.9& 48.5  & 24.0& 541.3    & 92.3       & 2.5       & 1.3      \\
Mamba-2& 63.0 & 44.6    & 51.7     & 55.9& 47.8  & 24.2& 24.9     & 91.9       & 20.6      & 2.3      \\
mLSTM  & 63.7 & 38.4    & 49.2     & 52.7& 41.7  & 27.7& 40.6     & 75.7       & 6 & 3.3      \\
Griffin& 63.7 & 43.3    & 51.5     & 52.8& 45.8  & 23.7& 93.0     & 89.1       & 17.8      & 0\\
HybridMamba    & 62.6 & 45.4    & 53.2     & 56.8& 49.6  & 23.9& 77.6     & 94.8       & 2 & 0\\ \bottomrule
\end{tabular}
}
\caption{Model performance on standardized benchmarks.}
    \label{tab:downstream_Task}
\end{table*}

%% file: figures/pg19_bin_scatter.tex
\begin{figure*}[!h]
    \centering
    \includegraphics[width=\linewidth]{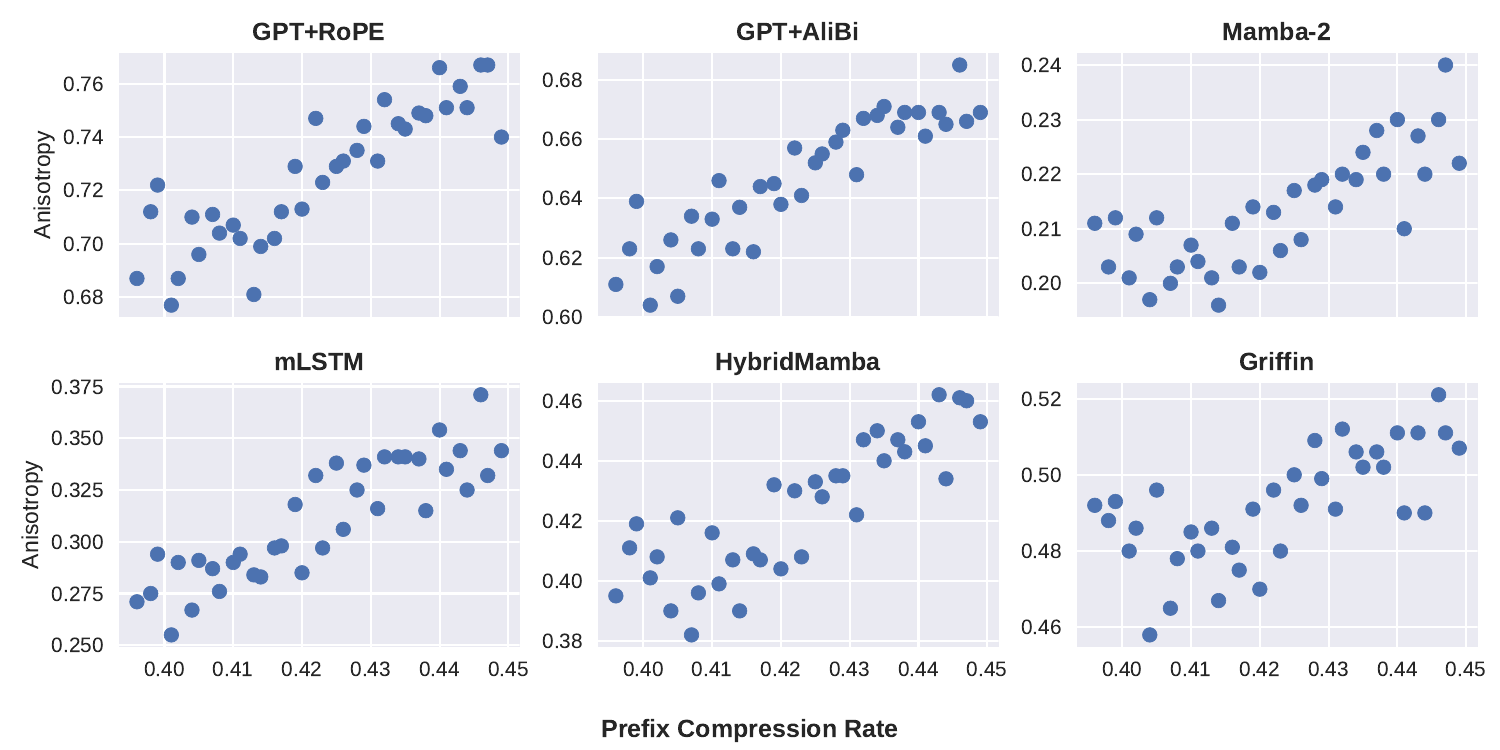}
    \caption{The increase in anisotropy as the compression rate increases is also observed when the language patterns are not disturbed, and across various architectures.}
    \label{fig:pg19_bin}
\end{figure*}

%% file: figures/niah_llama31.tex
\begin{figure}
    \centering
    \includegraphics[width=0.5\linewidth]{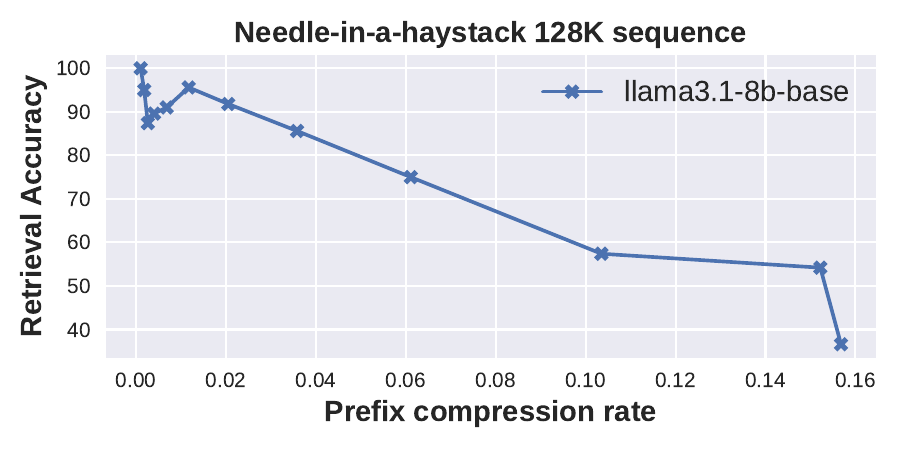}
    \caption{We insert increasing number of needles (key-value pair) into repetitive sentences (the ``haystack''). The number of needles vary from 2 to $\sim$7000. Inserting more needles each containing unique information makes the prefix less compressible, which happens alongside the increase in anisotropy. }
    \label{fig:niah_llama31}
\end{figure}

%% file: figures/collapse_representation.tex
\begin{figure*}[!h]
    \centering
    \includegraphics[width=0.5\linewidth]{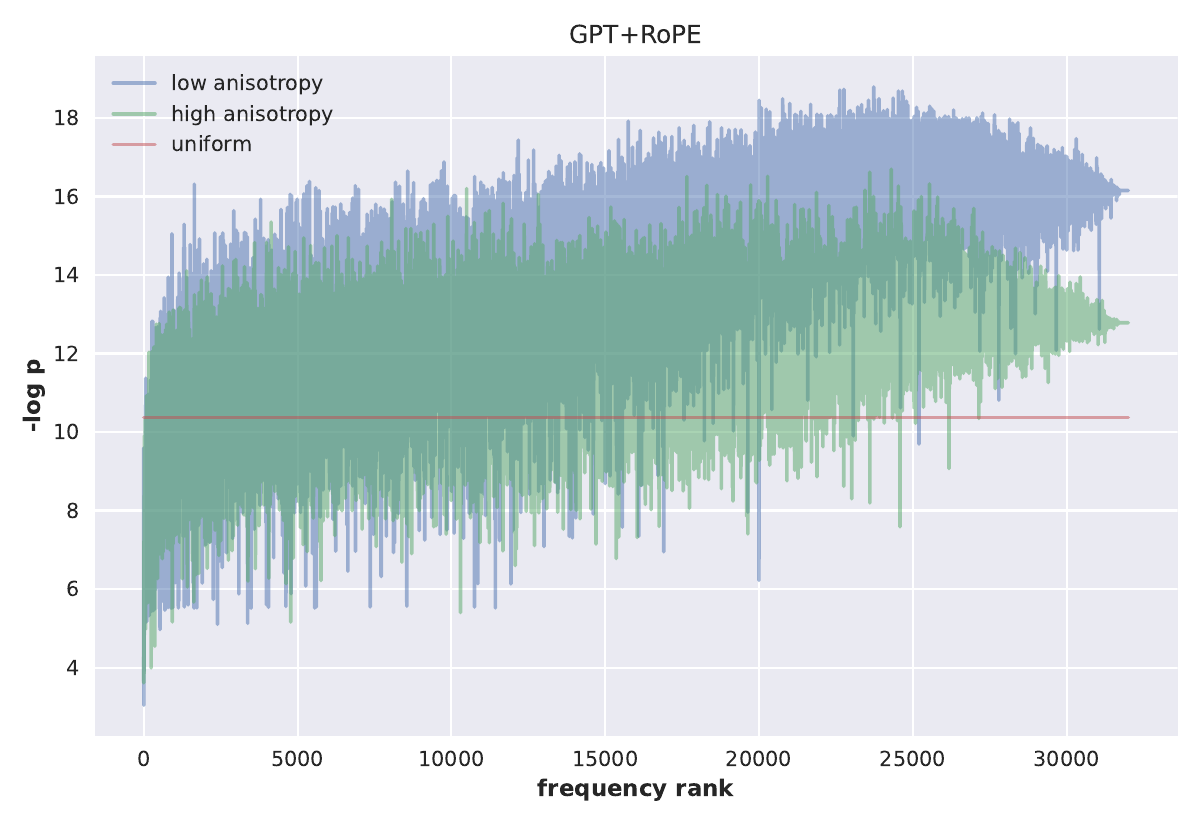}\includegraphics[width=0.5\linewidth]{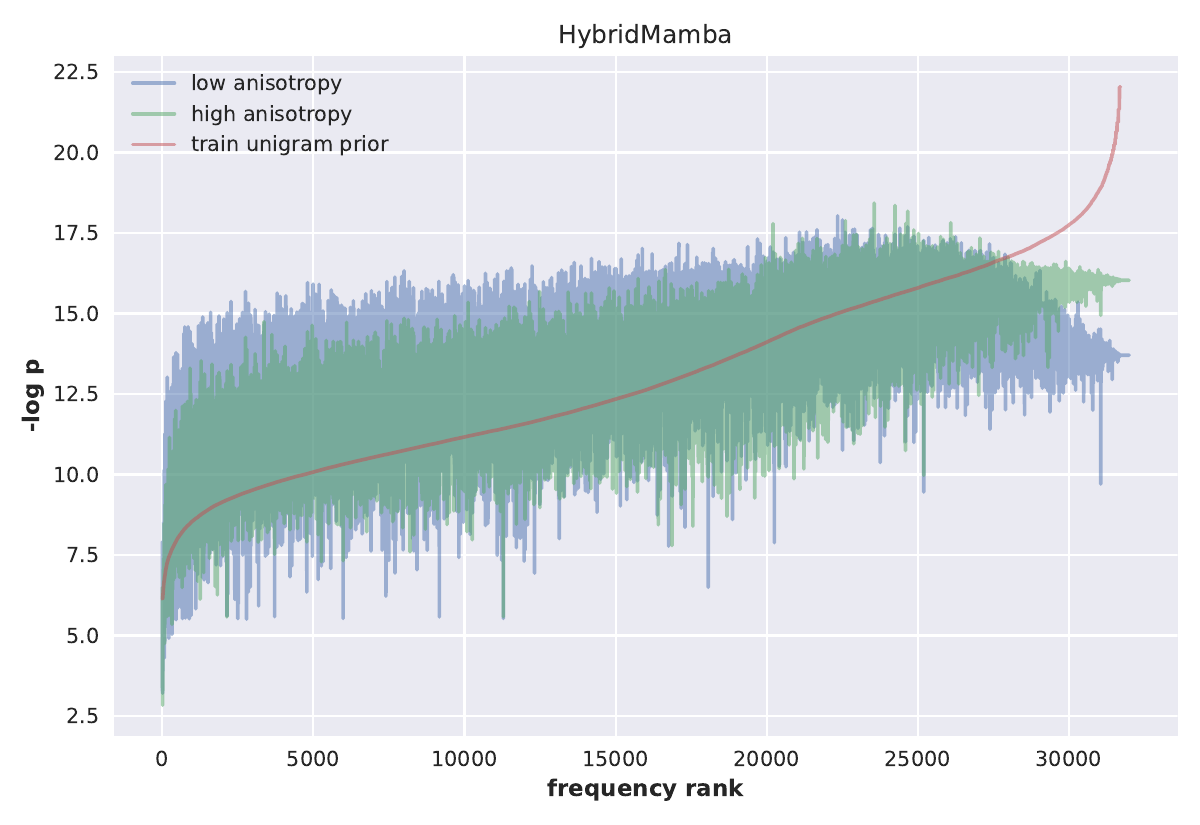}
    \caption{x-axis: token frequency rank based on pre-training data. y-axis: negative log probability over vocabulary average over all suffix tokens given various prefixes. (left) GPT+RoPE without any training-free extension. Low anisotropy: PG19 with 1K context length; high anisotropy: PG19 with 16K context length. (right) HybridMamba. Low anisotropy: PG19 with 1K context length; high anisotropy: PG19 with shuffled prefix, also with 1K context length.}
    \label{fig:collapse_repr}
\end{figure*}

%% file: tables/accs_pg19_slm.tex
\begin{table*}[!h]
    \centering
    \scalebox{0.55}{
    \begin{tabular}{@{}c|ccc|ccc|ccc|ccc|ccc|ccc@{}}
\toprule
\multicolumn{1}{l}{\multirow{2}{*}{\begin{tabular}[c]{@{}l@{}}Relative\\ Right Boundary\end{tabular}}} & \multicolumn{3}{c}{GPT+RoPE} & \multicolumn{3}{c}{GPT+ALiBi} & \multicolumn{3}{c}{Griffin} & \multicolumn{3}{c}{mLSTM} & \multicolumn{3}{c}{HybridMamba} & \multicolumn{3}{c}{Mamba-2} \\ \cmidrule(l){2-19} 
\multicolumn{1}{l}{}                                                                                   & A       & self-sim  & ACCS   & A       & self-sim   & ACCS   & A      & self-sim  & ACCS   & A      & self-sim & ACCS  & A        & self-sim   & ACCS    & A      & self-sim  & ACCS   \\ \midrule
0.1                                                                                                    & 0.932   & 0.972     & 0.039  & 0.640   & 1.000      & 0.360  & 0.677  & 0.952     & 0.275  & 0.272  & 1.000    & 0.727 & 0.595    & 0.975      & 0.380   & 0.202  & 1.000     & 0.798  \\
0.2                                                                                                    & 0.932   & 0.971     & 0.039  & 0.640   & 1.000      & 0.360  & 0.697  & 0.906     & 0.210  & 0.272  & 1.000    & 0.728 & 0.617    & 0.945      & 0.328   & 0.202  & 1.000     & 0.798  \\
0.3                                                                                                    & 0.932   & 0.970     & 0.037  & 0.640   & 1.000      & 0.360  & 0.703  & 0.886     & 0.183  & 0.271  & 0.999    & 0.728 & 0.627    & 0.925      & 0.298   & 0.202  & 1.000     & 0.798  \\
0.4                                                                                                    & 0.933   & 0.970     & 0.037  & 0.640   & 1.000      & 0.360  & 0.706  & 0.877     & 0.170  & 0.270  & 0.998    & 0.728 & 0.629    & 0.910      & 0.280   & 0.202  & 0.999     & 0.798  \\
0.5                                                                                                    & 0.933   & 0.969     & 0.036  & 0.640   & 1.000      & 0.360  & 0.709  & 0.875     & 0.167  & 0.268  & 0.996    & 0.728 & 0.636    & 0.891      & 0.255   & 0.202  & 0.999     & 0.798  \\
0.6                                                                                                    & 0.933   & 0.969     & 0.036  & 0.640   & 1.000      & 0.360  & 0.710  & 0.872     & 0.163  & 0.264  & 0.989    & 0.725 & 0.641    & 0.873      & 0.232   & 0.202  & 0.999     & 0.797  \\
0.7                                                                                                    & 0.933   & 0.969     & 0.035  & 0.640   & 1.000      & 0.360  & 0.710  & 0.871     & 0.161  & 0.259  & 0.977    & 0.718 & 0.643    & 0.866      & 0.223   & 0.203  & 0.998     & 0.796  \\
0.8                                                                                                    & 0.933   & 0.969     & 0.035  & 0.640   & 1.000      & 0.360  & 0.711  & 0.869     & 0.158  & 0.257  & 0.964    & 0.707 & 0.643    & 0.862      & 0.219   & 0.204  & 0.995     & 0.791  \\
0.9                                                                                                    & 0.933   & 0.969     & 0.035  & 0.640   & 1.000      & 0.360  & 0.711  & 0.865     & 0.154  & 0.263  & 0.954    & 0.691 & 0.645    & 0.859      & 0.214   & 0.209  & 0.981     & 0.772  \\
1.0                                                                                                    & 0.934   & 0.968     & 0.035  & 0.661   & 0.925      & 0.263  & 0.716  & 0.839     & 0.123  & 0.283  & 0.811    & 0.527 & 0.662    & 0.824      & 0.163   & 0.227  & 0.836     & 0.609  \\ \bottomrule
\end{tabular}
    }
    \caption{Anisotropy and self-similarity for calculating ACCS of models in Figure~\ref{fig:accs_vs_mdl} left.}
    \label{tab:accs_pg19_slm}
\end{table*}

%% file: tables/accs_pg19_llama.tex
\begin{table*}[!h]
    \centering
    \scalebox{0.65}{
    \begin{tabular}{@{}c|ccc|ccc|ccc|ccc@{}}
\toprule
\multicolumn{1}{l}{\multirow{2}{*}{\begin{tabular}[c]{@{}l@{}}Relative \\ Right Boundary\end{tabular}}} & \multicolumn{3}{c}{\texttt{llama3-8b-base}} & \multicolumn{3}{c}{\texttt{llama31-8b-base}} & \multicolumn{3}{c}{\texttt{llama31-8b-instruct}} & \multicolumn{3}{c}{\texttt{llama31-70b-base}} \\ \cmidrule(l){2-13} 
\multicolumn{1}{l}{}                                                                                    & A          & self-sim    & ACCS      & A          & self-sim     & ACCS      & A           & self-sim      & ACCS        & A            & self-sim      & ACCS        \\ \midrule
0.1                                                                                                     & 0.4841     & 0.8669      & 0.3829    & 0.3094     & 0.9971       & 0.6877    & 0.3458      & 0.9967        & 0.6509      & 0.1845       & 0.9955        & 0.8110      \\
0.2                                                                                                     & 0.482      & 0.7813      & 0.2994    & 0.3091     & 0.9953       & 0.6862    & 0.3452      & 0.9946        & 0.6494      & 0.1847       & 0.9935        & 0.8087      \\
0.3                                                                                                     & 0.481      & 0.7330      & 0.2515    & 0.3088     & 0.9934       & 0.6846    & 0.3448      & 0.9922        & 0.6474      & 0.1848       & 0.9912        & 0.8064      \\
0.4                                                                                                     & 0.485      & 0.6508      & 0.1657    & 0.3087     & 0.9912       & 0.6825    & 0.3447      & 0.9895        & 0.6448      & 0.1850       & 0.9888        & 0.8038      \\
0.5                                                                                                     & 0.487      & 0.6354      & 0.1483    & 0.3083     & 0.9889       & 0.6806    & 0.3439      & 0.9869        & 0.6430      & 0.1850       & 0.9858        & 0.8009      \\
0.6                                                                                                     & 0.488      & 0.6342      & 0.1466    & 0.3081     & 0.9862       & 0.6781    & 0.3435      & 0.9835        & 0.6400      & 0.1853       & 0.9825        & 0.7972      \\
0.7                                                                                                     & 0.487      & 0.6323      & 0.1449    & 0.3077     & 0.9827       & 0.6749    & 0.3422      & 0.9791        & 0.6369      & 0.1854       & 0.9779        & 0.7925      \\
0.8                                                                                                     & 0.489      & 0.6305      & 0.1419    & 0.3076     & 0.9776       & 0.6700    & 0.3414      & 0.9727        & 0.6313      & 0.1855       & 0.9713        & 0.7858      \\
0.9                                                                                                     & 0.488      & 0.6300      & 0.1418    & 0.3074     & 0.9686       & 0.6612    & 0.3400      & 0.9608        & 0.6208      & 0.1863       & 0.9595        & 0.7733      \\
1.0                                                                                                     & 0.488      & 0.6289      & 0.1412    & 0.3247     & 0.8490       & 0.5243    & 0.3608      & 0.8272        & 0.4664      & 0.2016       & 0.8037        & 0.6021      \\ \bottomrule
\end{tabular}
    }
    \caption{Anisotropy and self-similarity for calculating ACCS of models in Figure~\ref{fig:accs_vs_mdl} right.}
    \label{tab:accs_pg19_llama}
\end{table*}

%% file: tables/accs_synth_slm.tex
\begin{table*}[!h]
    \centering
    \scalebox{0.55}{
    \begin{tabular}{@{}c|ccc|ccc|ccc|ccc|ccc|ccc@{}}
\toprule
\multicolumn{1}{l}{\multirow{2}{*}{\begin{tabular}[c]{@{}l@{}}Sequence\\ Length\end{tabular}}} & \multicolumn{3}{c}{GPT+RoPE} & \multicolumn{3}{c}{GPT+ALiBi} & \multicolumn{3}{c}{Griffin} & \multicolumn{3}{c}{mLSTM} & \multicolumn{3}{c}{HybridMamba} & \multicolumn{3}{c}{Mamba-2} \\ \cmidrule(l){2-19} 
\multicolumn{1}{l}{}                                                                           & A      & self-sim  & ACCS   & A       & self-sim  & ACCS   & A      & self-sim  & ACCS   & A      & self-sim & ACCS  & A      & self-sim & ACCS  & A        & self-sim   & ACCS    \\ \midrule
1024                                                                                           & 0.624  & 0.931     & 0.307  & 0.622   & 0.924     & 0.302  & 0.695  & 0.955     & 0.260  & 0.407  & 0.909    & 0.501 & 0.598  & 0.911    & 0.312 & 0.552    & 0.901      & 0.350   \\
2048                                                                                           & 0.707  & 0.666     & -0.040 & 0.746   & 0.959     & 0.213  & 0.725  & 0.931     & 0.206  & 0.410  & 0.881    & 0.471 & 0.677  & 0.911    & 0.234 & 0.507    & 0.902      & 0.395   \\
4096                                                                                           & 0.801  & 0.777     & -0.024 & 0.807   & 0.969     & 0.162  & 0.704  & 0.905     & 0.202  & 0.404  & 0.892    & 0.488 & 0.710  & 0.933    & 0.223 & 0.479    & 0.871      & 0.392   \\
8192                                                                                           & 0.913  & 0.932     & 0.020  & 0.804   & 0.917     & 0.112  & 0.613  & 0.728     & 0.116  & 0.394  & 0.720    & 0.325 & 0.764  & 0.893    & 0.129 & 0.478    & 0.737      & 0.260   \\
12288                                                                                          & 0.909  & 0.927     & 0.018  & 0.798   & 0.881     & 0.084  & 0.630  & 0.707     & 0.077  & 0.386  & 0.613    & 0.227 & 0.771  & 0.861    & 0.091 & 0.489    & 0.662      & 0.173   \\
16384                                                                                          & 0.957  & 0.968     & 0.011  & 0.797   & 0.865     & 0.068  & 0.623  & 0.684     & 0.061  & 0.388  & 0.562    & 0.174 & 0.798  & 0.865    & 0.067 & 0.497    & 0.626      & 0.128   \\ \bottomrule
\end{tabular}
    }
    \caption{Anisotropy and self-similarity for calculating ACCS of models in Figure~\ref{fig:accs_vs_mdl_varlen} left.}
    \label{tab:accs_synth_slm}
\end{table*}

%% file: tables/accs_synth_llama.tex
\begin{table*}[]
    \centering
    \scalebox{0.65}{
    \begin{tabular}{@{}c|ccc|ccc|ccc|ccc@{}}
\toprule
\multicolumn{1}{l}{\multirow{2}{*}{\begin{tabular}[c]{@{}l@{}}Sequence\\ Length\end{tabular}}} & \multicolumn{3}{c}{\texttt{llama3-8b-base}} & \multicolumn{3}{c}{\texttt{llama31-8b-base}} & \multicolumn{3}{c}{\texttt{llama31-8b-instruct}} & \multicolumn{3}{c}{\texttt{llama31-70b-base}} \\ \cmidrule(l){2-13} 
\multicolumn{1}{l}{}                                                                           & A         & self-sim     & ACCS      & A          & self-sim     & ACCS      & A           & self-sim       & ACCS       & A           & self-sim       & ACCS        \\ \midrule
8192                                                                                           & 0.543     & 0.672        & 0.129     & 0.500      & 0.697        & 0.197     & 0.581       & 0.745          & 0.164      & 0.392       & 0.619          & 0.227       \\
10240                                                                                          & 0.634     & 0.784        & 0.150     & 0.443      & 0.623        & 0.180     & 0.489       & 0.661          & 0.173      & 0.331       & 0.539          & 0.208       \\
13312                                                                                          & 0.630     & 0.705        & 0.075     & 0.408      & 0.557        & 0.149     & 0.455       & 0.588          & 0.133      & 0.290       & 0.453          & 0.163       \\
16384                                                                                          & 0.610     & 0.639        & 0.029     & 0.389      & 0.520        & 0.131     & 0.421       & 0.544          & 0.123      & 0.266       & 0.407          & 0.142       \\
20480                                                                                          & 0.666     & 0.643        & -0.023    & 0.380      & 0.477        & 0.096     & 0.417       & 0.498          & 0.081      & 0.255       & 0.356          & 0.101       \\ \bottomrule
\end{tabular}
    }
    \caption{Anisotropy and self-similarity for calculating ACCS of models in Figure~\ref{fig:accs_vs_mdl_varlen} right.}
    \label{tab:accs_synth_llama}
\end{table*}

%% file: tables/rope_base_table.tex
\begin{table*}[]
    \centering
    \scalebox{0.65}{
    \begin{tabular}{@{}rcccccccccccc@{}}
    \toprule
\multicolumn{1}{l}{}                         & \multicolumn{4}{c}{PPL \& suffix PPL (Intrinsic)}                    & \multicolumn{5}{c}{RULER S-NIAH (extrinsic)} & \multicolumn{3}{c}{Representation-level Eval} \\ \midrule
\multicolumn{1}{l}{\begin{tabular}[c]{@{}l@{}}ROPE\\ BASE\end{tabular}} & \begin{tabular}[c]{@{}c@{}}16K, suffix\\ (token level)\end{tabular} & \begin{tabular}[c]{@{}c@{}}16K\\ (word level)\end{tabular} & \begin{tabular}[c]{@{}c@{}}4K, suffix\\ (token level)\end{tabular} & \begin{tabular}[c]{@{}c@{}}4K\\ (word level)\end{tabular} & 1K      & 2K      & 4K      & 8K     & Avg.  & self-similarity    & Anisotropy    & ACCS     \\ \midrule
50000             & 2761.6        & 88398.2                         & 801.1        & 753.3                          & 0.8933  & 0.7653  & 0.0000  & 0.0000 & 0.415 & 0.8209             & 0.6955        & 0.1254   \\
54000             & 2883.7        & 79676.7                         & 661.3        & 614.9                          & 0.8807  & 0.7493  & 0.0000  & 0.0000 & 0.408 & 0.8349             & 0.6955        & 0.1393   \\
58000             & 2854.6        & 72999.0                         & 505.3        & 515.9                          & 0.8707  & 0.7053  & 0.0000  & 0.0000 & 0.394 & 0.8558             & 0.6942        & 0.1616   \\
62000             & 3046.0        & 65074.9                         & 415.6        & 456.2                          & 0.8580  & 0.6667  & 0.0007  & 0.0000 & 0.381 & 0.8737             & 0.6868        & 0.1870   \\
65000             & 3182.5        & 61059.9                         & 366.6        & 419.3                          & 0.8407  & 0.6480  & 0.0193  & 0.0000 & 0.377 & 0.8824             & 0.6772        & 0.2052   \\
66000             & 3228.2        & 59490.9                         & 325.1        & 397.7                          & 0.8340  & 0.6520  & 0.0560  & 0.0000 & 0.386 & 0.8868             & 0.6673        & 0.2195   \\
67000             & 3122.5        & 57985.7                         & 331.3        & 391.3                          & 0.8273  & 0.6467  & 0.0727  & 0.0000 & 0.387 & 0.8902             & 0.6689        & 0.2213   \\
69000             & 3213.7        & 55498.7                         & 259.9        & 365.8                          & 0.8307  & 0.6493  & 0.0940  & 0.0000 & 0.394 & 0.9024             & 0.6593        & 0.2431   \\
70000             & 3197.1        & 54445.4                         & 222.2        & 357.0                          & 0.8267  & 0.6447  & 0.1040  & 0.0000 & 0.394 & 0.9084             & 0.6550        & 0.2534   \\
72000             & 3183.5        & 52174.7                         & 150.0        & 340.2                          & 0.8280  & 0.6487  & 0.1107  & 0.0000 & 0.397 & 0.9164             & 0.6392        & 0.2772   \\
73000             & 3158.8        & 50687.4                         & 122.8        & 335.4                          & 0.8260  & 0.6467  & 0.1180  & 0.0000 & 0.398 & 0.9205             & 0.6309        & 0.2896   \\
75000             & 3010.0        & 48012.0                         & 90.5         & 326.3                          & 0.8240  & 0.6440  & 0.1107  & 0.0000 & 0.395 & 0.9233             & 0.6141        & 0.3092   \\
77000             & 2837.1        & 45972.1                         & 73.5         & 320.2                          & 0.8213  & 0.6380  & 0.0880  & 0.0000 & 0.387 & 0.9239             & 0.5962        & 0.3277   \\
80000             & 2694.1        & 43650.6                         & 62.7         & 314.7                          & 0.8167  & 0.6287  & 0.1220  & 0.0000 & 0.392 & 0.9218             & 0.5801        & 0.3417   \\
81000             & 2685.4        & 42992.8                         & 58.3         & 313.2                          & 0.8113  & 0.6227  & 0.1507  & 0.0000 & 0.396 & 0.9217             & 0.5715        & 0.3502   \\
82000             & 2687.9        & 41995.8                         & 55.8         & 312.5                          & 0.8140  & 0.6180  & 0.2153  & 0.0000 & 0.412 & 0.9208             & 0.5655        & 0.3553   \\
83000             & 2707.2        & 41358.2                         & 52.7         & 311.9                          & 0.8107  & 0.6120  & 0.2747  & 0.0000 & 0.424 & 0.9214             & 0.5591        & 0.3624   \\
84000             & 2707.4        & 40457.6                         & 51.0         & 311.3                          & 0.8087  & 0.6027  & 0.3080  & 0.0000 & 0.430 & 0.9208             & 0.5537        & 0.3671   \\
85000             & 2739.0        & 39837.3                         & 48.6         & 310.9                          & 0.8027  & 0.5987  & 0.3093  & 0.0000 & 0.428 & 0.9209             & 0.5472        & 0.3737   \\
90000             & 2755.2        & 35929.9                         & 43.5         & 312.3                          & 0.7840  & 0.5713  & 0.3707  & 0.0000 & 0.432 & 0.9146             & 0.5233        & 0.3913   \\
92000             & 2740.5        & 34629.3                         & 42.9         & 313.1                          & 0.7800  & 0.5627  & 0.3607  & 0.0000 & 0.426 & 0.9128             & 0.5163        & 0.3965   \\
94000             & 2708.5        & 33359.1                         & 42.2         & 313.2                          & 0.7680  & 0.5500  & 0.3413  & 0.0000 & 0.415 & 0.9124             & 0.5101        & 0.4022   \\
96000             & 2608.0        & 32199.1                         & 41.3         & 313.7                          & 0.7653  & 0.5393  & 0.3227  & 0.0000 & 0.407 & 0.9137             & 0.5055        & 0.4082   \\
98000             & 2481.6        & 31087.2                         & 40.2         & 314.5                          & 0.7520  & 0.5193  & 0.2940  & 0.0000 & 0.391 & 0.9173             & 0.5015        & 0.4158   \\
100000            & 2373.2        & 29981.0                         & 39.3         & 315.6                          & 0.7480  & 0.5140  & 0.2760  & 0.0000 & 0.385 & 0.9213             & 0.4983        & 0.4229   \\
104000            & 2253.8        & 28031.2                         & 37.9         & 319.0                          & 0.7387  & 0.4973  & 0.2433  & 0.0000 & 0.370 & 0.9293             & 0.4932        & 0.4361   \\
108000            & 2284.1        & 26465.7                         & 36.9         & 322.9                          & 0.7233  & 0.4807  & 0.2140  & 0.0000 & 0.355 & 0.9355             & 0.4901        & 0.4455   \\
112000            & 2287.5        & 24894.3                         & 36.4         & 326.6                          & 0.7093  & 0.4600  & 0.2020  & 0.0000 & 0.343 & 0.9392             & 0.4888        & 0.4505   \\
116000            & 2222.2        & 23617.6                         & 36.2         & 331.7                          & 0.6940  & 0.4467  & 0.2053  & 0.0000 & 0.337 & 0.9418             & 0.4891        & 0.4527   \\
120000            & 2166.3        & 22285.0                         & 36.2         & 338.1                          & 0.6780  & 0.4227  & 0.2093  & 0.0000 & 0.328 & 0.9440             & 0.4902        & 0.4538   \\
140000            & 1888.7        & 17380.1                         & 37.1         & 375.2                          & 0.6207  & 0.3347  & 0.1800  & 0.0000 & 0.284 & 0.9591             & 0.4940        & 0.4651   \\
160000            & 1463.0        & 14503.1                         & 40.5         & 423.9                          & 0.5733  & 0.3320  & 0.1987  & 0.0000 & 0.276 & 0.9682             & 0.5018        & 0.4664   \\
180000            & 1569.1        & 12871.9                         & 46.4         & 501.8                          & 0.4907  & 0.2820  & 0.1693  & 0.0000 & 0.236 & 0.9702             & 0.5139        & 0.4563   \\
200000            & 1615.1        & 12710.0                         & 61.9         & 722.9                          & 0.3620  & 0.1647  & 0.1353  & 0.0000 & 0.166 & 0.9677             & 0.5371        & 0.4306   \\
220000            & 1464.2        & 10418.5                         & 62.1         & 693.3                          & 0.4333  & 0.2153  & 0.1840  & 0.0000 & 0.208 & 0.9669             & 0.5331        & 0.4338   \\
260000            & 1308.0        & 8841.7                          & 74.2         & 877.2                          & 0.3987  & 0.1600  & 0.1100  & 0.0000 & 0.167 & 0.9638             & 0.5488        & 0.4150   \\
280000            & 1237.1        & 8272.2                          & 79.2         & 983.1                          & 0.3800  & 0.1553  & 0.0840  & 0.0033 & 0.156 & 0.9620             & 0.5566        & 0.4054   \\
300000            & 1198.8        & 7636.3                          & 82.8         & 1059.2                         & 0.3553  & 0.1407  & 0.0807  & 0.0680 & 0.161 & 0.9605             & 0.5634        & 0.3971   \\
320000            & 1131.0        & 7157.5                          & 88.7         & 1151.9                         & 0.3327  & 0.1233  & 0.0567  & 0.0347 & 0.137 & 0.9594             & 0.5693        & 0.3902   \\
330000            & 1101.7        & 6943.7                          & 90.6         & 1190.4                         & 0.3147  & 0.1153  & 0.0493  & 0.0260 & 0.126 & 0.9593             & 0.5715        & 0.3879   \\
340000            & 1073.9        & 6808.6                          & 93.8         & 1245.4                         & 0.3040  & 0.1080  & 0.0440  & 0.0173 & 0.118 & 0.9590             & 0.5753        & 0.3837   \\
360000            & 993.0         & 6450.8                          & 98.3         & 1330.0                         & 0.2880  & 0.1007  & 0.0367  & 0.0167 & 0.111 & 0.9585             & 0.5800        & 0.3785   \\
380000            & 942.3         & 6086.2                          & 101.8        & 1401.6                         & 0.2820  & 0.0887  & 0.0340  & 0.0107 & 0.104 & 0.9582             & 0.5842        & 0.3740   \\
400000            & 899.9         & 5896.0                          & 107.0        & 1503.4                         & 0.2413  & 0.0740  & 0.0187  & 0.0087 & 0.086 & 0.9582             & 0.5895        & 0.3687   \\
450000            & 791.0         & 5555.0                          & 119.9        & 1751.2                         & 0.1900  & 0.0467  & 0.0107  & 0.0053 & 0.063 & 0.9585             & 0.6020        & 0.3565   \\
500000            & 538.0         & 5121.7                          & 127.5        & 1912.6                         & 0.1707  & 0.0393  & 0.0073  & 0.0027 & 0.055 & 0.9588             & 0.6107        & 0.3481   \\
600000            & 391.2         & 5206.4                          & 145.2        & 2319.4                         & 0.0900  & 0.0140  & 0.0007  & 0.0013 & 0.027 & 0.9596             & 0.6256        & 0.3340   \\
800000            & 313.7         & 6089.0                          & 175.7        & 3067.6                         & 0.0420  & 0.0000  & 0.0000  & 0.0000 & 0.011 & 0.9589             & 0.6481        & 0.3108   \\
900000            & 327.8         & 6557.6                          & 187.2        & 3411.9                         & 0.0300  & 0.0000  & 0.0000  & 0.0000 & 0.008 & 0.9585             & 0.6552        & 0.3034   \\
1000000           & 336.8         & 6892.8                          & 195.0        & 3655.0                         & 0.0207  & 0.0000  & 0.0000  & 0.0000 & 0.005 & 0.9591             & 0.6593        & 0.2998  \\
\bottomrule
\end{tabular}
    }
    \caption{Intrinsic, extrinsic, and representation-level evaluation while varying RoPE base $\theta$. Descriptions can be found in section~\ref{sec:case_transformer}. We compute suffix perplexity on the token-level, while reporting whole-chunk perplexity on word-level, as implemented in lm\_eval\_harness.}
    \label{tab:rope_base_table}
\end{table*}